\documentclass{article}

\usepackage[utf8]{inputenc}
\usepackage{amsmath}
\usepackage{caption}
\usepackage{subcaption}

\usepackage{authblk}
\usepackage[square,sort,comma,numbers]{natbib}
\usepackage{graphicx}

\providecommand{\keywords}[1]
{
  \small	
  \textbf{\textit{Keywords---}} #1
}

\begin{document}

\title{Learning velocity model for complex media with deep convolutional neural networks}
\author[1]{A.Stankevich}
\author[1]{I.Nechepurenko}
\author[1]{A.Shevchenko}
\author[2]{L.Gremyachikh}
\author[2,1]{A.Ustyuzhanin}
\author[1]{A.Vasyukov}

\affil[1]{Moscow Institute of Physics and Technology, Russia}
\affil[2]{HSE University, Russia}

\maketitle

\begin{abstract}
The paper considers the problem of velocity model acquisition for a complex media based on boundary measurements. The acoustic model is used to describe the media. We used an open-source dataset of velocity distributions to compare the presented results with the previous works directly. Forward modeling is performed using the grid-characteristic numerical method. The inverse problem is solved using deep convolutional neural networks. Modifications for a baseline UNet architecture are proposed to improve both structural similarity index measure quantitative correspondence of the velocity profiles with the ground truth. We evaluate our enhancements and demonstrate the statistical significance of the results.
\end{abstract}


\section{Introduction\label{sec:intro}}

The problem of identifying elastic media properties based on their measured response is a well-known one. This problem has many applications and variations in industrial non-destructive testing, seismic exploration, biomedical engineering, and other areas.
This paper considers methods based on acoustic or elastic wave excitation in a media under consideration, recording the media's response and identifying the media's properties from this response. This problem statement is typical for ultrasonic techniques and seismic imaging.

There are many different approaches for solving an inverse problem to determine the spatial distribution of mechanical properties from the recorded response. New methods have emerged recently based on the success in deep convolutional neural networks research and development. The media's response is used as an input for the neural network that predicts the required properties. The feasibility of this approach was demonstrated for different problems.

The works \cite{Ye2018, Tripathi2019} successfully applied deep learning for non-destructive testing data interpretation and defect classification. The approaches presented can work in real-time and provide new options compared with traditional imaging and interpretation methods.

Several cases demonstrate the convolutional network's feasibility for different biomedical imaging problems. The paper~\cite{Perdios2020} studied an application of deep learning to artifacts correction on single-shot ultrasound images obtained with sparse linear arrays. The initial results demonstrated an image quality comparable or better to that obtained from conventional beamforming. The work~\cite{Stankevich2021} used neural networks to identify the shape of the aberration prism that distorts the ultrasonic signal. The work~\cite{PATEL2019} applied convolutional networks to elasticity imaging to distinguish benign tumors from their malignant counterparts based on measured displacement fields on the boundary of the domain.

This paper contributed to the same area. We present the results primarily for seismic use cases, but this approach, together with models, methods, and a numerical pipeline, can be easily generalized and transferred to other problem domains.

The success of deep convolutional neural networks for image processing caused a significant interest in applying them to the problem of seismic segmentation and interpretation.
The works \cite{Waldeland2017, Waldeland2018} presented an approach to discriminate between salt and other geological structures based on the seismic response. Related results were presented in \cite{shi2018, zhao2018} and many other works.

This initial success of deep convolutional neural networks in seismic interpretation caused complementary works that contributed to open datasets. The paper \cite{Alaudah2019} covered the facies classification problem and published an open-source, fully annotated 3D geologic model of a genuine Dutch F3 block. The work \cite{baroni2019} contributed Penobscot -- another open dataset for seismic interpretation problems.

These datasets became de-facto benchmarks for comparing different machine-learning approaches for facies classification. The availability of these benchmarks empowered further research in this area. It caused an open-source domain-specific framework \cite{Salvaris2020} that applies state-of-the-art segmentation algorithms for seismic interpretation.

Similar ideas are also applied to seismic impedance inversion and acquiring velocity model of a media. Novel methods complement classical approaches for seismic inversion problems. The techniques based on neural networks don't require initial guesses and are extremely fast. These features make them attractive for a quick assessment of large datasets.

The works \cite{Liu2019, Alfarraj2019, Wang2020} presented a promising approach based on two complementary neural networks that allowed the fast adaptation of the baseline model for the site under investigation using transfer learning techniques. The authors demonstrated the high quality of the predicted results. The approach relies on a cleaned dataset and is fine-tuned exclusively for seismic exploration cases when drilling is possible.

Another approach was taken in \cite{Yang2019, Das2019, Dujardin2020, Araya-Polo2019, Park2020}. These works consider boundary observations only. This technique is more demanding than the previous one from an inverse problem point of view. However, it is preferable from the engineering point of view for a wide range of use cases when drilling is complicated. Moreover, this approach can be generalized to other areas besides seismic exploration, where invasive data acquisition is impossible.

Our work contributes to the area of velocity model acquisition for the complex heterogeneous media-based exclusively on boundary measurements. We present several enhancements for the baseline neural networks and demonstrate them using an open-source dataset of 1600 velocity models shown in \cite{Yang2019}. We believe that this will also contribute to establishing domain-specific benchmark datasets and empower further research in this area.

\section{Model and method} 

\subsection{Numerical pipeline}

\begin{figure}[!h]
    \begin{subfigure}[b]{1.0\textwidth}
        \begin{subfigure}[b]{0.5\textwidth}
           \centering
           \includegraphics[width=\textwidth]{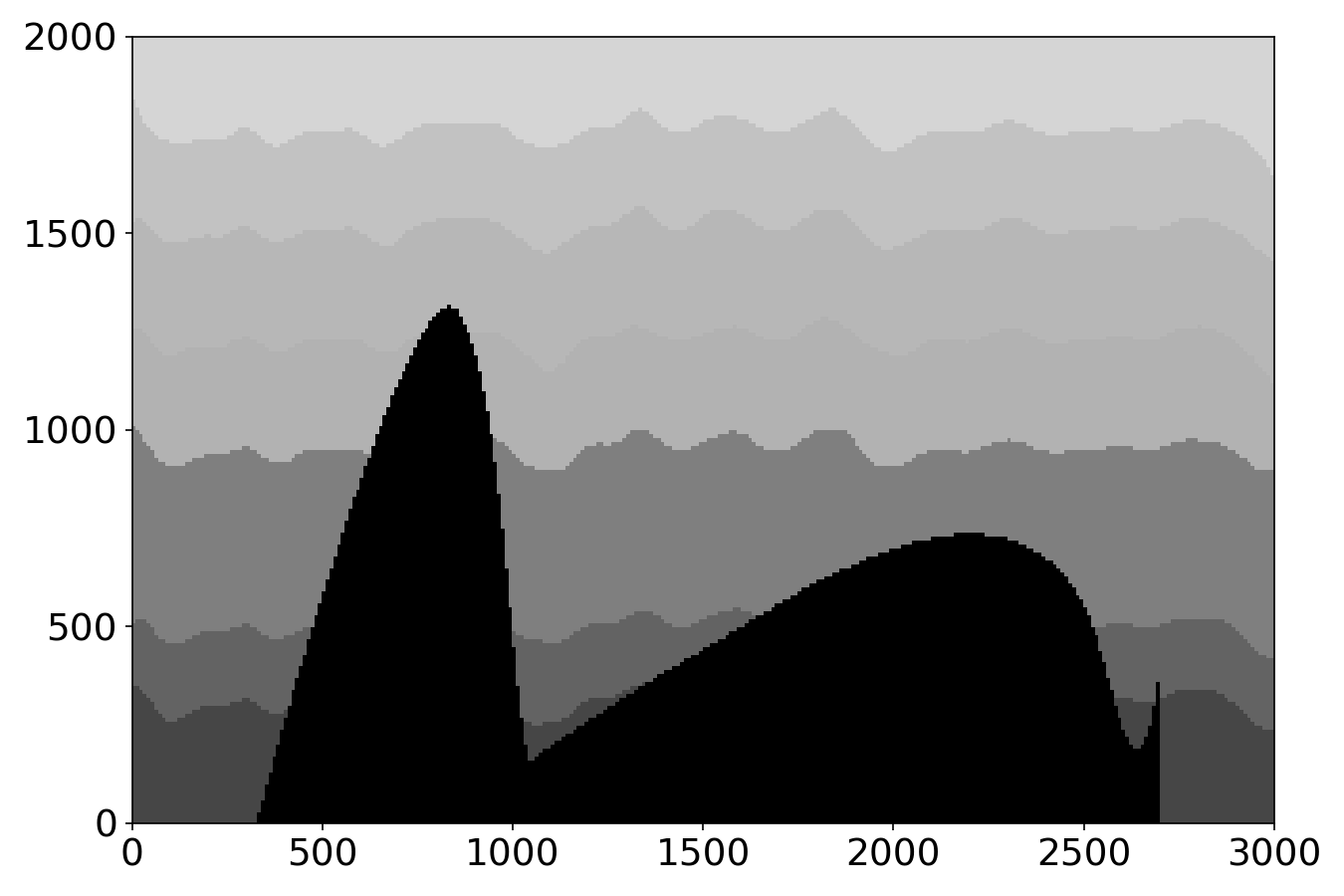}
           \caption{Harder inclusion on the bottom}
           \label{fig:sample_velocity_model_1}
        \end{subfigure}
        \begin{subfigure}[b]{0.5\textwidth}
            \centering
            \includegraphics[width=\textwidth]{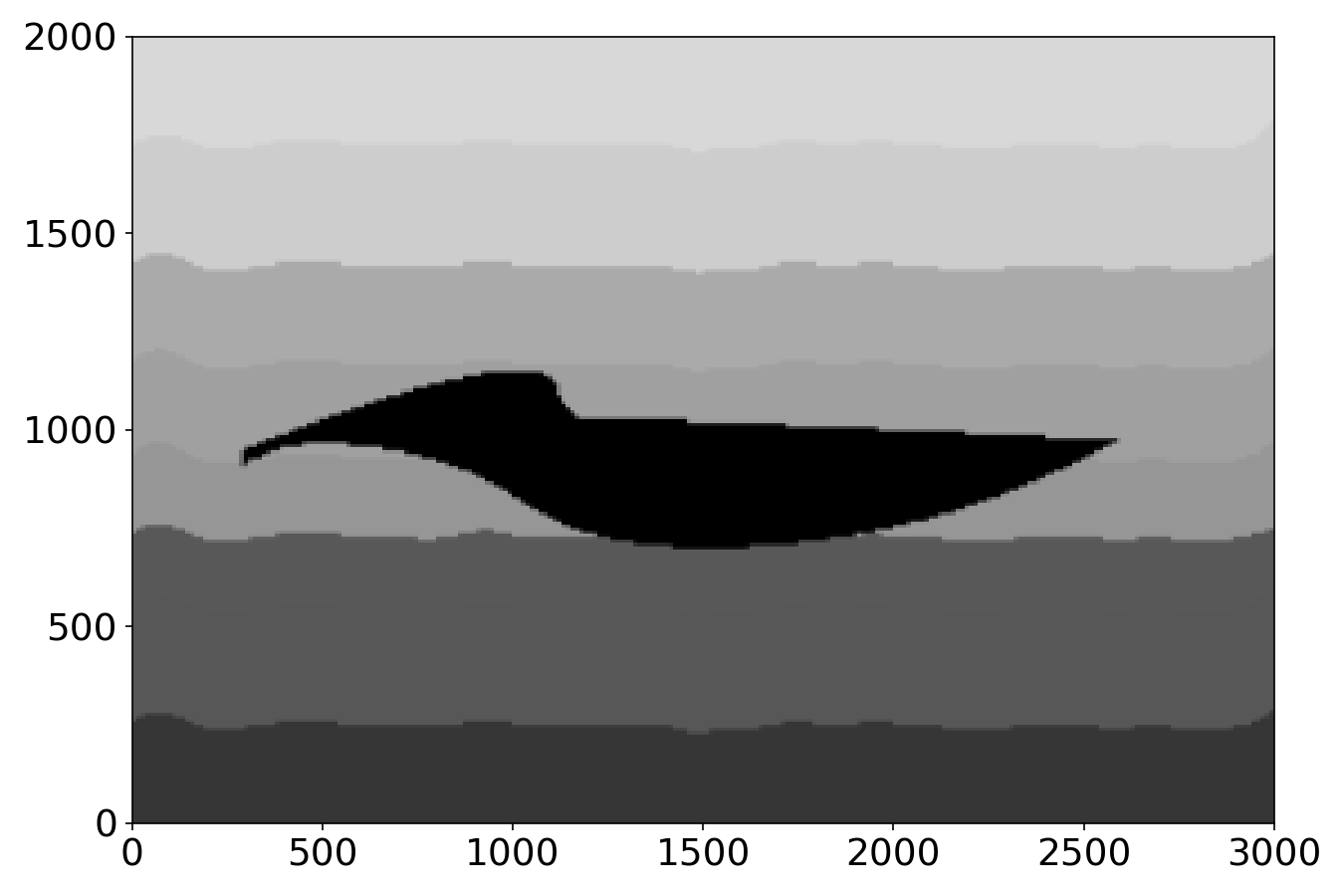}
            \caption{Harder inclusion in the middle}
            \label{fig:sample_velocity_model_2}
        \end{subfigure}
    \end{subfigure}
\caption{Sample velocity models}
\label{fig:sample_velocity_model}
\end{figure}

The dataset from \cite{Yang2019} is used in this paper. Each sample has a length along the OX axis of 3000 meters and a depth of 2000 meters along the OY axis. Sound speed in this dataset varies typically from 2500 m/s in softer media to 4500 m/s in harder media. Figure \ref{fig:sample_velocity_model} presents velocity models for 2 of 1600 samples. The Black area is an acoustically hard zone located in relatively soft background media.

The seismic response was calculated numerically for each velocity distribution as described in the section \ref{subsec:forward_modeling}. Thus we created 1600 pairs of ground truth velocity models and corresponding seismic responses recorded on the domain's boundary. 
The data was split into training, validational, and testing subsets using random sampling with 70\%/15\%/15\% split ratio. The training data sample was fed to a deep convolutional neural network described in the section \ref{subsec:inverse_problem}. The network trains to predict velocity distribution from the seismic response. The quality of the predictions was evaluated on the testing dataset.

\subsection{Forward modeling}\label{subsec:forward_modeling}

Direct problem uses an acoustic model following the velocity models in the dataset. The media is described with the following equations:
\begin{align}
    \label{initial_equation_acoustic}
	\rho\dot{\vec{v}} &= - \nabla p   \nonumber\\
	\dot{p} &= -K (\nabla \cdot \vec{v}). 
\end{align}

In these equations \(p \) -
pressure, \( \vec{v} \) - velocity vector,  \( \rho \) - medium density, \( K \) - bulk modulus.

Forward modeling uses grid-characteristic numerical method, that demonstrated good quality of the numerical solution for various wave dynamics problems, including wave propagation in complex composite structures \cite{gcm_ac, gcm_residual}, classical seismic migration \cite{gcm_migration}, wave propagation in a fractured \cite{gcm_fractured} and multicomponent \cite{gcm_multicomponent} geological media.

The condition of the free boundary was specified at the upper boundary of the computational domain. An absorbing boundary condition was used on all the other borders.
The emitter of the scanning pulse was moved along the upper boundary of the domain, and 9 equidistant emitter positions were used for each velocity model sample. There were 300 equidistant receivers on the upper boundary of the domain to record the response.

\begin{figure}[!h]
    \begin{subfigure}[b]{1.0\textwidth}
        \begin{subfigure}[b]{0.3\textwidth}
           \centering
           \includegraphics[width=\textwidth]{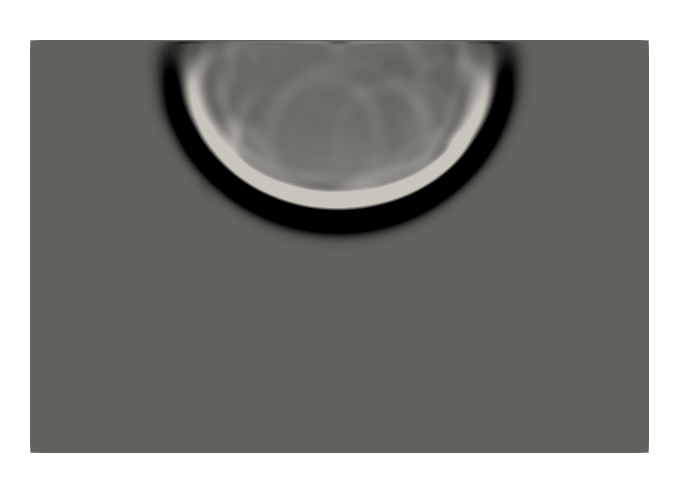}
           \caption{\(t = 0.3s\)}
        \end{subfigure}
        \begin{subfigure}[b]{0.3\textwidth}
            \centering
            \includegraphics[width=\textwidth]{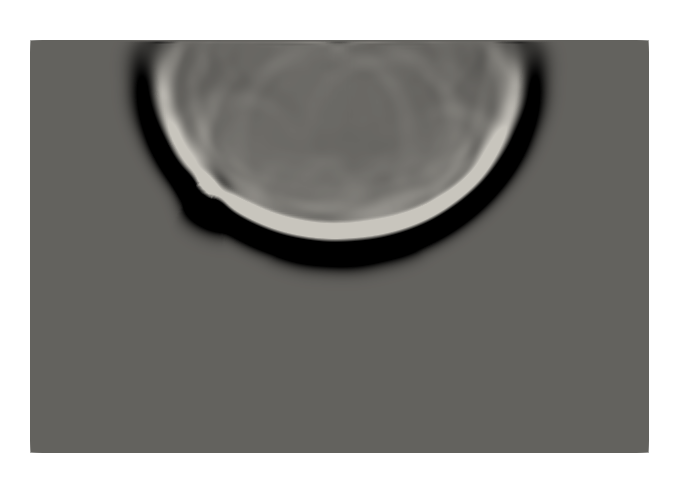}
            \caption{\(t = 0.35s\)}
        \end{subfigure}
        \begin{subfigure}[b]{0.3\textwidth}
            \centering
            \includegraphics[width=\textwidth]{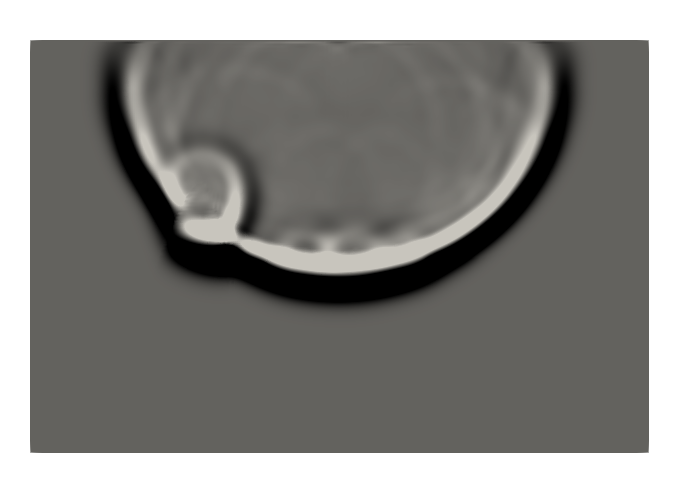}
            \caption{\(t = 0.4s\)}
        \end{subfigure}
    \end{subfigure}
    \begin{subfigure}[b]{1.0\textwidth}
        \begin{subfigure}[b]{0.3\textwidth}
           \centering
           \includegraphics[width=\textwidth]{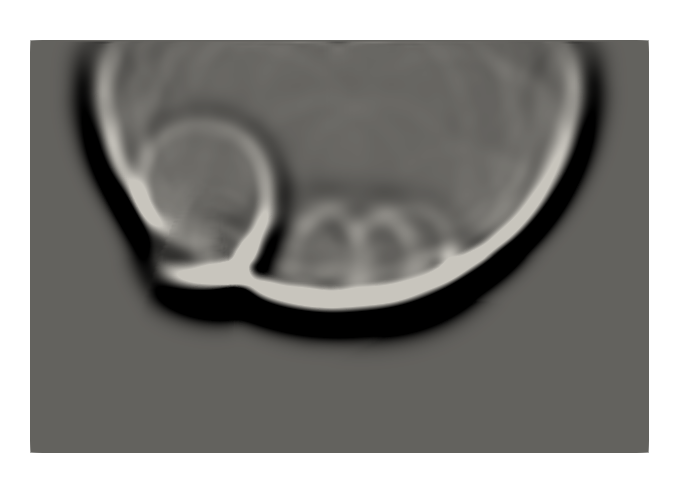}
           \caption{\(t = 0.45s\)}
        \end{subfigure}
        \begin{subfigure}[b]{0.3\textwidth}
            \centering
            \includegraphics[width=\textwidth]{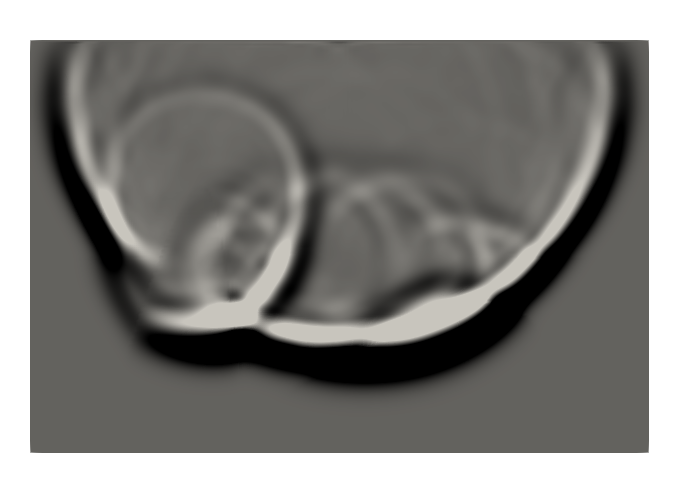}
            \caption{\(t = 0.5s\)}
        \end{subfigure}
        \begin{subfigure}[b]{0.3\textwidth}
            \centering
            \includegraphics[width=\textwidth]{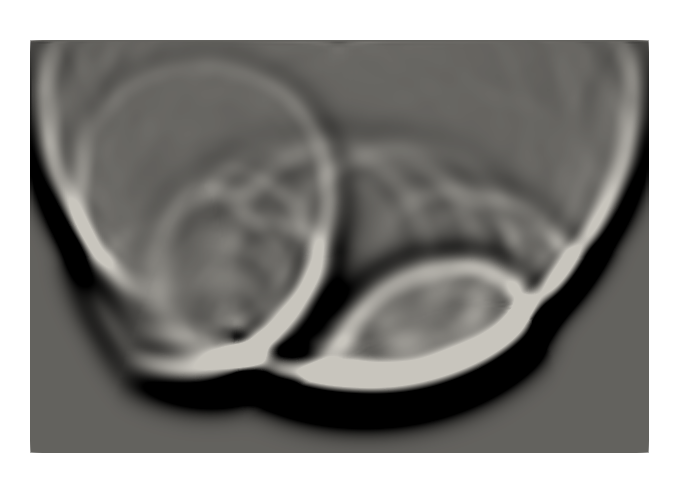}
            \caption{\(t = 0.55s\)}
        \end{subfigure}
    \end{subfigure}
    \caption{Sample wave propagation, emitter located in the center of the domain}
    \label{fig:direct_model_sample_pos50}
\end{figure}

\begin{figure}[!h]
    \begin{subfigure}[b]{1.0\textwidth}
        \begin{subfigure}[b]{0.3\textwidth}
           \centering
           \includegraphics[width=\textwidth]{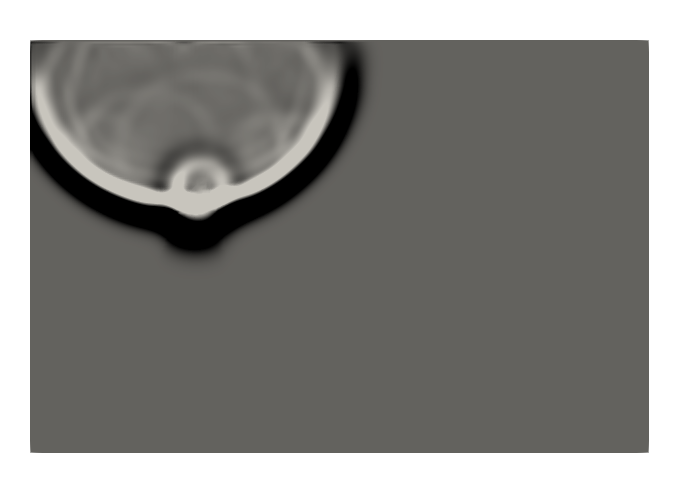}
           \caption{\(t = 0.3s\)}
        \end{subfigure}
        \begin{subfigure}[b]{0.3\textwidth}
            \centering
            \includegraphics[width=\textwidth]{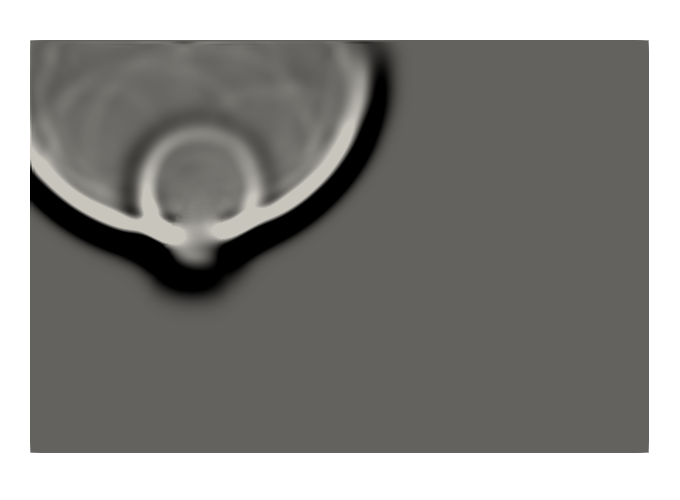}
            \caption{\(t = 0.35s\)}
        \end{subfigure}
        \begin{subfigure}[b]{0.3\textwidth}
            \centering
            \includegraphics[width=\textwidth]{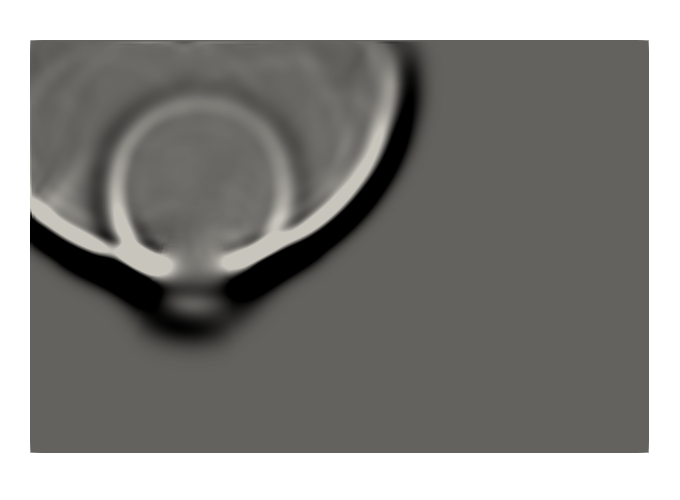}
            \caption{\(t = 0.4s\)}
        \end{subfigure}
    \end{subfigure}
    \begin{subfigure}[b]{1.0\textwidth}
        \begin{subfigure}[b]{0.3\textwidth}
           \centering
           \includegraphics[width=\textwidth]{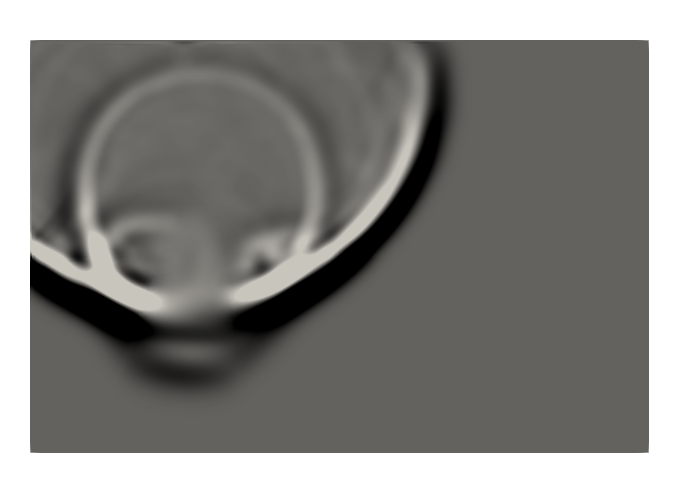}
           \caption{\(t = 0.45s\)}
        \end{subfigure}
        \begin{subfigure}[b]{0.3\textwidth}
            \centering
            \includegraphics[width=\textwidth]{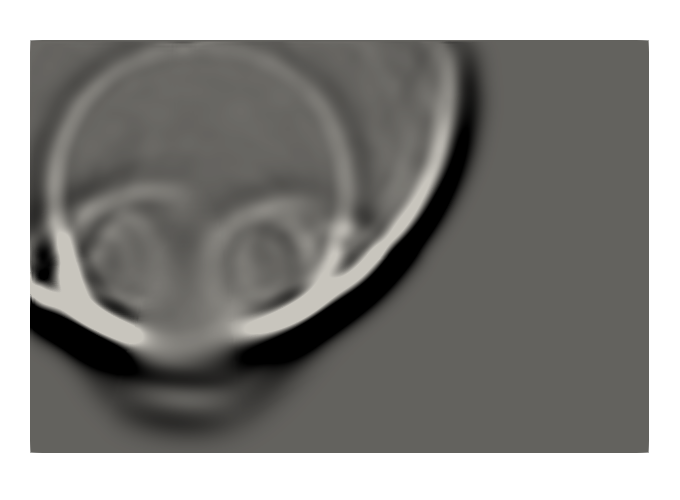}
            \caption{\(t = 0.5s\)}
        \end{subfigure}
        \begin{subfigure}[b]{0.3\textwidth}
            \centering
            \includegraphics[width=\textwidth]{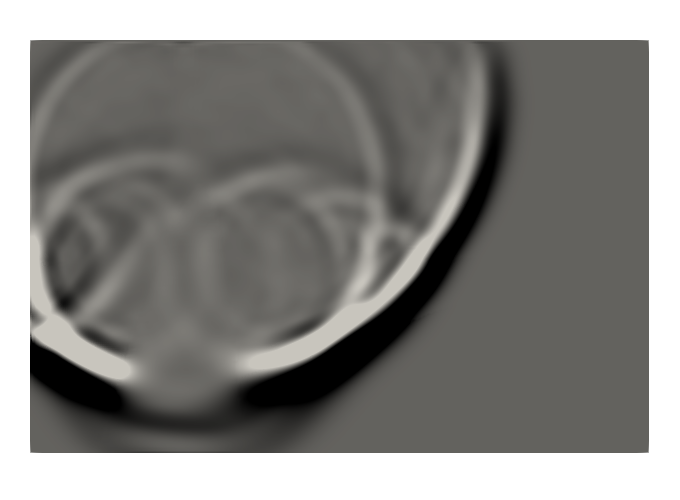}
            \caption{\(t = 0.55s\)}
        \end{subfigure}
    \end{subfigure}
    \caption{Sample wave propagation, emitter located in the left part of the domain}
    \label{fig:direct_model_sample_pos25}
\end{figure}

Figure \ref{fig:direct_model_sample_pos50} presents an example of forward modeling results. The media has the velocity distribution as presented in Figure \ref{fig:sample_velocity_model_1}, the modeling covers 2 seconds of physical time. One can see that during an initial phase of scanning pulse propagation (\(t = 0.3s\)), there are only low amplitude responses from the low contrast boundaries between the layers of the background media. The scanning pulse reaches the left peak of the hard object at the moment \(t = 0.35s\), and the first high amplitude response forms. A medium amplitude response from the medium contrast boundary in the background media happens later (\(t = 0.45s\)). The high amplitude response from the right peak of the hard object starts to form around \(t = 0.5s\). Figure \ref{fig:direct_model_sample_pos25} shows the wave patterns in the media at the same moments for a different position of the emitter.


These single-shot responses for the different positions of the emitter for the same velocity distribution are stacked together and form a complete multi-shot seismic response data for the media under investigation.

The dataset produced by forward modeling was published as \cite{numerical_dataset}. We believe the availability of these data will enable community contribution to future research.

\subsection{Inverse problem}\label{subsec:inverse_problem}

Following previous research, we utilize UNet \cite{ronneberger2015unet}, a fully convolutional encoder-decoder neural network architecture, to predict seismic velocity distribution. Originally UNet was designed to solve problems of semantic image segmentation, taking RGB image as input and producing segmentation mask as an output. Our adaptation assumes stacking rescaled seismograms, obtained from different source positions, to form an N-channeled image to feed into UNet. The number of channels equals the number of emitter positions. The network output, respectively, is interpreted as the velocity distribution in the media. 

Our work additionally contributes three modifications to the default pipeline of solving such problems, increasing its overall precision.

First, we suggest adding the real- and complex-valued part of the 2D Fourier images of seismograms as additional channels to the input of the neural network. Different seismogram sources contribute as separate channels to the input tensor.

Second, we experimented with analogs for classical regularization techniques. In classical methods for solving inverse problems, the uniqueness of the solution is often achieved by adding a regularizing term to the loss function, which endows the solution with additional properties, such as smoothness. To achieve a similar effect in application to neural network training, we suggest the regularization, obtained by calculating the $L_2$ norm of the result of the application of Sobel filter \cite{sobel_filter} to the output of the network. The impact of the Sobel filter is similar to that of the operator of numerical differentiation. Thus, we expect the obtained solutions to have a smoother structure than those without regularization.

Third, we combined several instances of the neural network into an ensemble by averaging their outputs. Ensembling is a well-known technique for various machine learning problems. In this context, ensembling serves as another analog for classical regularization to obtain a smoother structure of the predicted velocity model.

\section{Numerical Experiments}

\subsection{Data representation} 

The dataset $\mathcal{D}$ consists of pairs $(d_i, s_i)$, where $s_i$ is the seismic response from the velocity distribution $d_i$, calculated as presented in Section \ref{subsec:forward_modeling}. Velocity distributions are represented with $300 \times 200 $ real-valued arrays, corresponding to values of $c_p$ in nodes of a regular mesh, covering the computational domain. In turn, seismograms are described with $300 \times 200 \times n $ real-valued arrays, containing vertical velocity component of the upper boundary of the domain recorded at 100 Hz frequency. Here $n$ is the total number of shots (emitter positions) that make up a complete seismic response for the sample. We additionally rescale seismograms to match 0. - 1. range.  

\subsection{Neural network training and evaluation}

We train all the models for a maximum of 50 epochs using Adam optimization algorithm \cite{kingma2014adam} with constant learning rate 1e-4, using mini-batches of size 10. The training was performed using the MSE loss function.

Due to the general deficiency of literature on machine learning applications to geological inversion, there is no precedence for using the exact metrics for the quality assessment of neural network predictions. We utilize structural similarity index measure (SSIM) \cite{SSIM} to evaluate the performance of model instances since the same metric was mentioned in \cite{Yang2019} for the experiments with the same dataset.
Hence, the choice of the resulting neural network weights was made by the epoch when it reaches a maximum value of SSIM metric \cite{SSIM} on the validational data fold. The whole training pipeline is implemented using PyTorch deep learning framework \cite{pytorch}.

To study the variance of model predictions depending on data, we utilize bootstrapping procedure to resample training subsets. The training procedure for the neural network was repeated 10 times for each suggested modification of the network. So, each new model instance was trained with a different training subset and demonstrated different results. For the rest of this section, we measure network performance on a testing sample after it was trained on a training and validation samples.

\subsection{Ablation study}

The following ablation study demonstrates the causations in the proposed architecture. We compare the quality of the solution by varying or disabling the pipeline steps one by one. This test identifies the parts of the setup that are essential for the quality of the solution and unnecessary features that are not statistically significant.

\begin{figure}[!h]
    \begin{subfigure}[b]{1.0\textwidth}
        \begin{subfigure}[b]{0.32\textwidth}
           \centering
           \includegraphics[width=\textwidth]{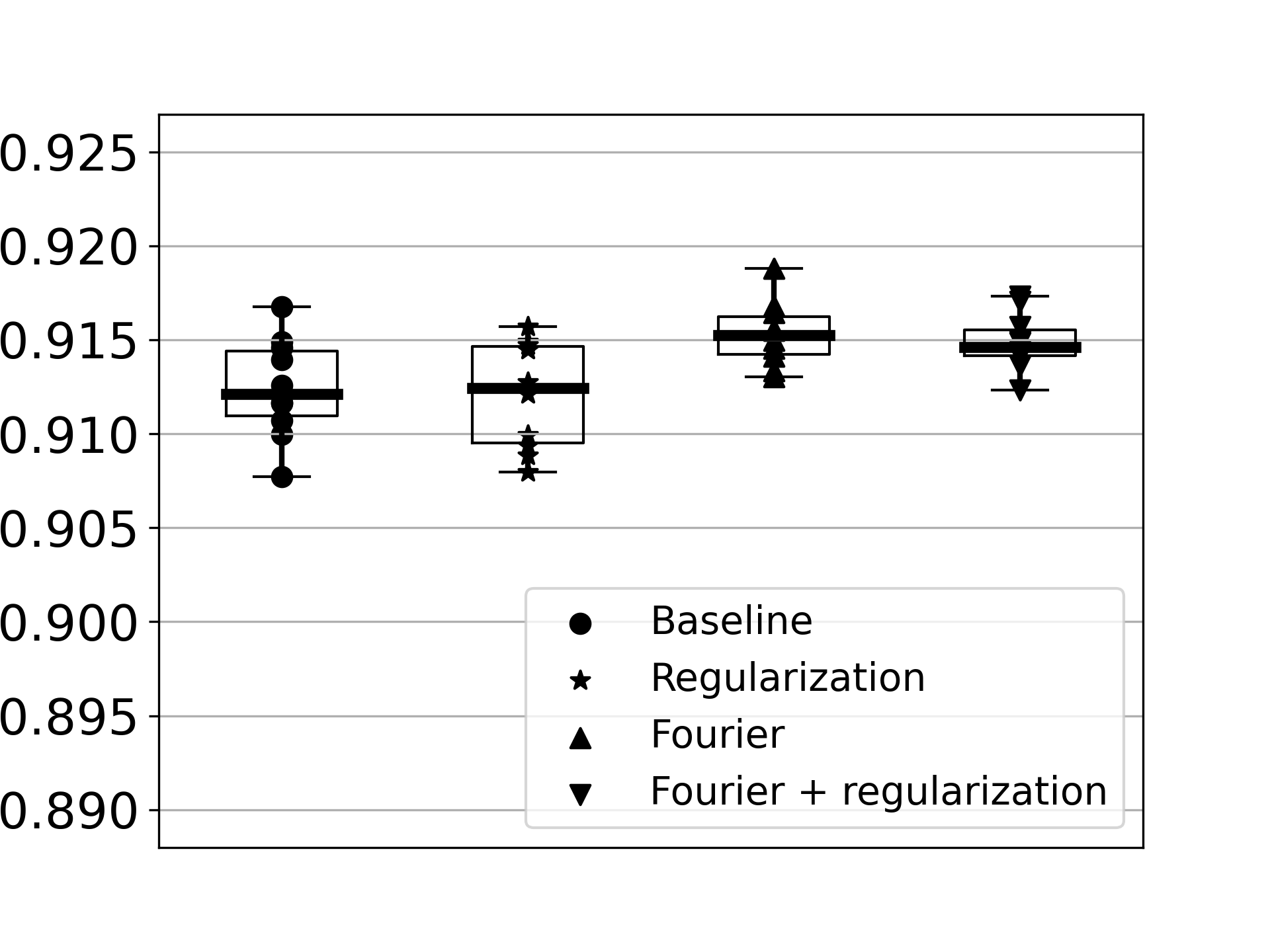}
           \caption{Single-shot data}
        \end{subfigure}
        \begin{subfigure}[b]{0.32\textwidth}
            \centering
            \includegraphics[width=\textwidth]{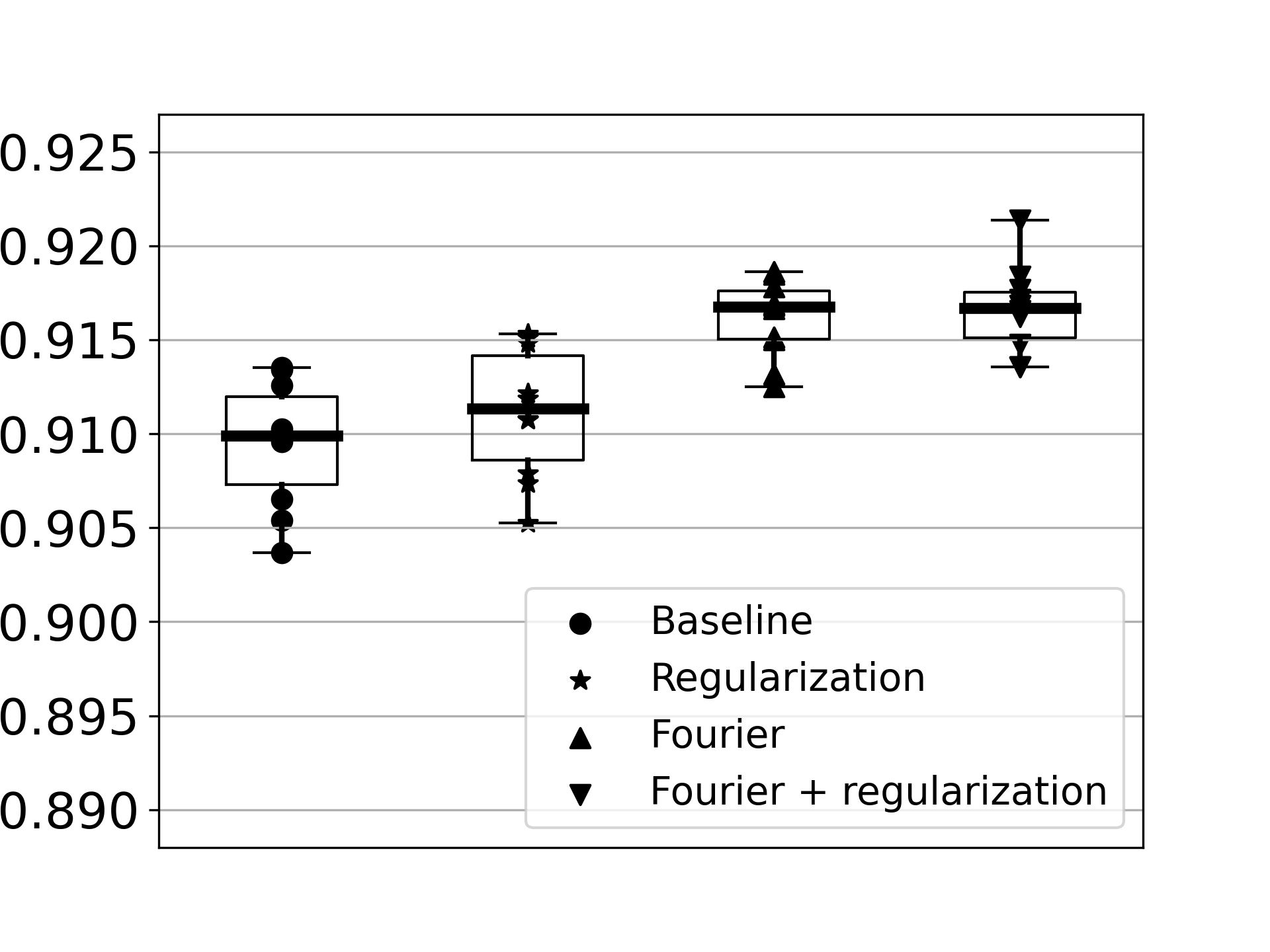}
            \caption{Triple-shot data}
        \end{subfigure}
        \begin{subfigure}[b]{0.32\textwidth}
            \centering
            \includegraphics[width=\textwidth]{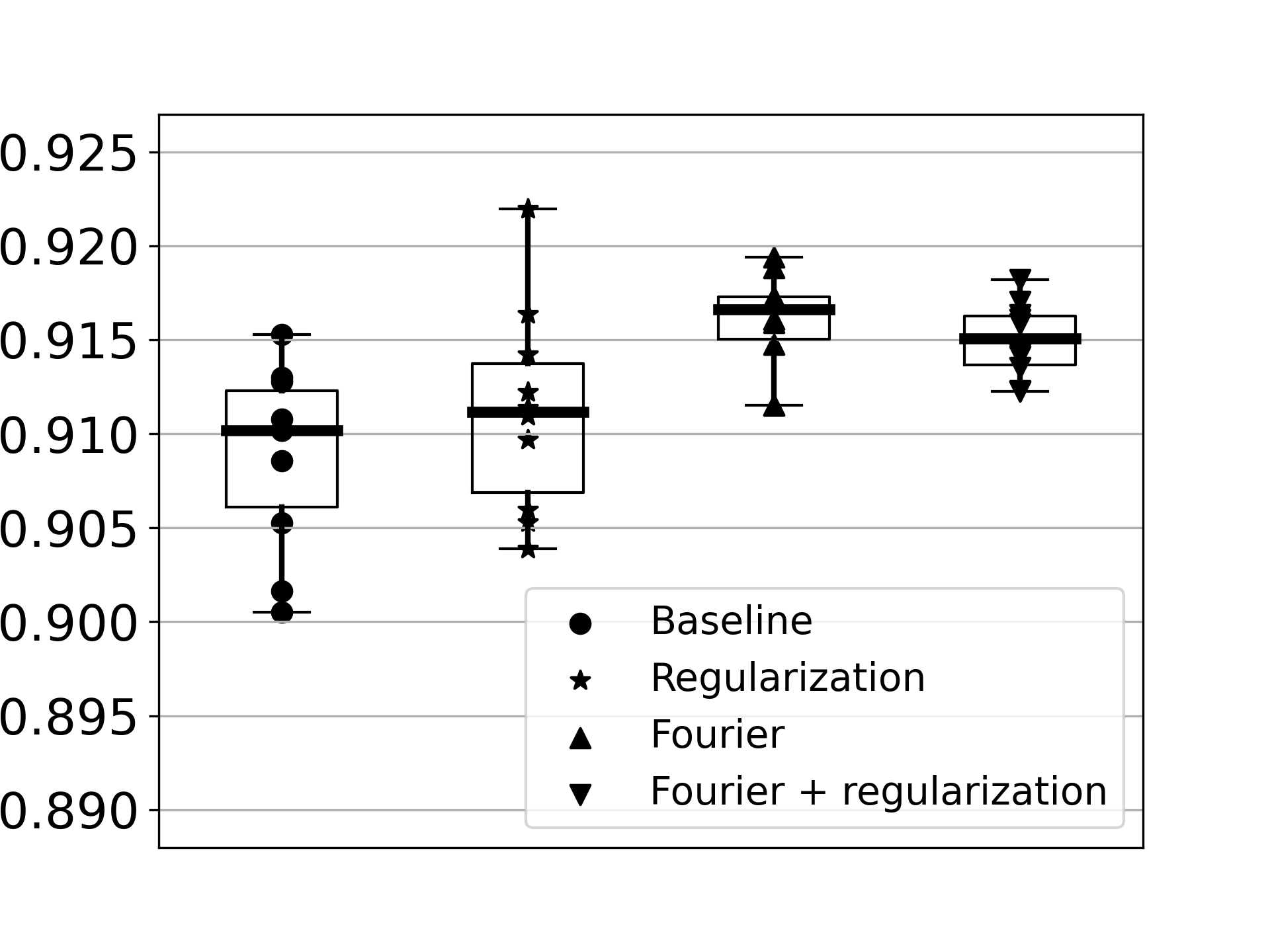}
            \caption{Nine-shot data}
        \end{subfigure}
    \end{subfigure}
    
    \caption{Distributions of SSIM values for bootstrapped model training}
    \label{fig:results_bootstrap_MSE}
    
\end{figure}

Table \ref{table:summatry_SSIM_MSE} summarizes experiments conducted with Fourier images and proposed regularization. The table provides average SSIM values for models trained on multishot seismograms. Distributions of SSIM values for model trained using different bootstrapped datasets on single-shot / triple-shot / nine-shot seismograms are depicted by Figure \ref{fig:results_bootstrap_MSE}.
Aggregated data allows us to hypothesize that complementing input tensors with Fourier images of seismograms positively affects the quality of predictions. At the same time, the impact of adding a regularization term is negligible.  

We conduct a series of statistical tests on populations of SSIM test values to confirm this statement. First, we verify that SSIM values for models instances may be considered normally distributed, employing Wilk-Shapiro test \cite{shapiro_wilk_statistical_test} with 0.05 alpha-level (see Table \ref{table:p_values_Shapiro_Wilk}). Further, we examine the equality of populations variances with Levene's test, grouping populations by the number of seismogram shots (Table \ref{table:p_values_Levene}). As neither of the tests rejects the respective null hypotheses, we assume that model populations satisfy all the required criteria for ANOVA \cite{ANOVA} testing.

\begin{table}[!h]
\centering
\begin{tabular}{|c|c|c|c|}
\hline
\textbf{Number of shots} & \textbf{Fourier} & \textbf{Regularization} & \textbf{Average SSIM} \\ \hline
1               & False          & False          & $ 0.912 \pm 0.003 $ \\ \hline
1               & True           & False          & $ \mathbf{0.915} \pm \mathbf{0.002} $\\ \hline
1               & False          & True           & $ 0.912 \pm 0.003 $ \\ \hline
1               & True           & True           & $ 0.915 \pm 0.001 $ \\ \hline
3               & False          & False          & $ 0.910 \pm 0.003 $ \\ \hline
3               & True           & False          & $ 0.916 \pm 0.002 $ \\ \hline
3               & False          & True           & $ 0.911 \pm 0.003 $ \\ \hline
3               & True           & True           & $ \mathbf{0.917} \pm \mathbf{0.002} $\\ \hline
9               & False          & False          & $ 0.909 \pm 0.005 $\\ \hline
9               & True           & False          & $ \mathbf{0.916} \pm \mathbf{0.003} $ \\ \hline
9               & False          & True           & $ 0.911 \pm 0.004 $ \\ \hline
9               & True           & True           & $ 0.915 \pm 0.002 $ \\ \hline
\end{tabular}
\caption{Average SSIM for the test dataset for models trained with MSE criterion}
\label{table:summatry_SSIM_MSE}
\end{table}

\begin{table}[!h]
\centering
\begin{tabular}{|c|c|c|c|}
\hline
\textbf{Shots} & \textbf{Fourier} & \textbf{Regularization} & \textbf{p-value} \\ \hline
1               & False          & False          & 0.990 \\ \hline
1               & True           & False          & 0.254 \\ \hline
1               & False          & True           & 0.784 \\ \hline
1               & True           & True           & 0.712 \\ \hline
3               & False          & False          & 0.365 \\ \hline
3               & True           & False          & 0.439 \\ \hline
3               & False          & True           & 0.376 \\ \hline
3               & True           & True           & 0.545 \\ \hline
9               & False          & False          & 0.363 \\ \hline
9               & True           & False          & 0.771 \\ \hline
9               & False          & True           & 0.186 \\ \hline
9               & True           & True           & 0.647 \\ \hline
\end{tabular}
\caption{p-values for Shapiro-Wilk statistics for sets of model instances trained on bootstrapped data}
\label{table:p_values_Shapiro_Wilk}
\end{table}

\begin{table}[!h]
\centering
\begin{tabular}{|c|c|}
\hline
\textbf{Shots} & \textbf{p-value for Levene's test} \\ \hline
1     & 0.060                     \\ \hline
3     & 0.371                     \\ \hline
9     & 0.070                     \\ \hline
\end{tabular}
\caption{p-values for Levene's test for sets of model instances grouped by number of shots in seismograms}
\label{table:p_values_Levene}
\end{table}

Table \ref{table:p_values_ANOVA_regularization} demonstrates that the contribution of regularization from the point of view of increasing the value of SSIM metric is statistically negligible. On the contrary, ANOVA-testing rejects the hypothesis about the equality of population means for models trained with and without adding Fourier images of seismograms to input tensors (see \ref{table:p_values_ANOVA_fourier}). Thus, we conclude that additional information provided by Fourier images enhances the framework's performance on a statistically significant level.

\begin{table}[!h]
\centering
\begin{tabular}{|c|c|c|}
\hline
\textbf{Shots} & \textbf{Fourier} & \textbf{p-value for ANOVA statistic} \\ \hline
1     & False          & 0.770 \\ \hline
1     & True           & 0.541 \\ \hline
3     & False          & 0.291 \\ \hline
3     & True           & 0.630 \\ \hline
9     & False          & 0.321 \\ \hline
9     & True           & 0.375 \\ \hline
\end{tabular}
\caption{p-values for ANOVA statistic, calculated for model populations trained with and without regularization}
\label{table:p_values_ANOVA_regularization}
\end{table}

\begin{table}[!h]
\centering
\begin{tabular}{|c|c|c|}
\hline
\textbf{Shots} & \textbf{Regularization} & \textbf{p-value for ANOVA statistic} \\ \hline
1     & False          & 1.42e-2 \\ \hline
1     & True           & 1.01e-2 \\ \hline
3     & False          & 4.93e-5 \\ \hline
3     & True           & 6.02e-4 \\ \hline
9     & False          & 7.62e-4 \\ \hline
9     & True           & 4.80e-2 \\ \hline
\end{tabular}
\caption{p-values for ANOVA statistic, calculated for model populations trained with and without adding fourier images}
\label{table:p_values_ANOVA_fourier}
\end{table}

The research has shown that suggested neural network architecture has enough statistical capacity to achieve low bias on the training dataset. However, it turned out that regularization methods, presented earlier, cannot resolve the issue of relatively poor generalization of the model. We suggest combining model instances, trained before, into an ensemble, by averaging their outputs to overcome the issue. 

Table \ref{table:summary_SSIM_MSE_ensembles} showcases the value of MSE on test subset for ensembles, composed from model instances trained before. Compared to constituent models, such modification yields an additional $1.\%$ gain in the value of target metrics.

\begin{table}[!h]
\centering
\begin{tabular}{|c|c|c|c|c|}
\hline
 \textbf{Number of shots} & \textbf{Fourier} & \textbf{Regularization} & \textbf{SSIM}   \\ \hline
 1               & False          & False          & 0.929 \\ \hline
 1               & True           & False          & 0.928 \\ \hline
 1               & False          & True           & 0.929 \\ \hline
 1               & True           & True           & 0.928 \\ \hline
 3               & False          & False          & 0.926 \\ \hline
 3               & False          & True           & 0.927 \\ \hline
 3               & True           & False          & 0.930 \\ \hline
 3               & True           & True           & 0.929 \\ \hline
 9               & False          & False          & 0.927 \\ \hline
 9               & True           & False          & 0.931 \\ \hline
 9               & False          & True           & 0.928 \\ \hline
 9               & True           & True           & 0.930 \\ \hline
\end{tabular}
\caption{Average SSIM for the test dataset for ensembles of models trained with MSE criterion}
\label{table:summary_SSIM_MSE_ensembles}
\end{table}

The ablation study results showed that Fourier images and ensembles contribute to the quality of the solution, and the proposed regularization approach is not significant.

\subsection{Velocity distribution prediction results}

\begin{figure}[!h]
    \begin{subfigure}[b]{1.0\textwidth}
        \begin{subfigure}[b]{0.5\textwidth}
           \centering
           \includegraphics[width=\textwidth]{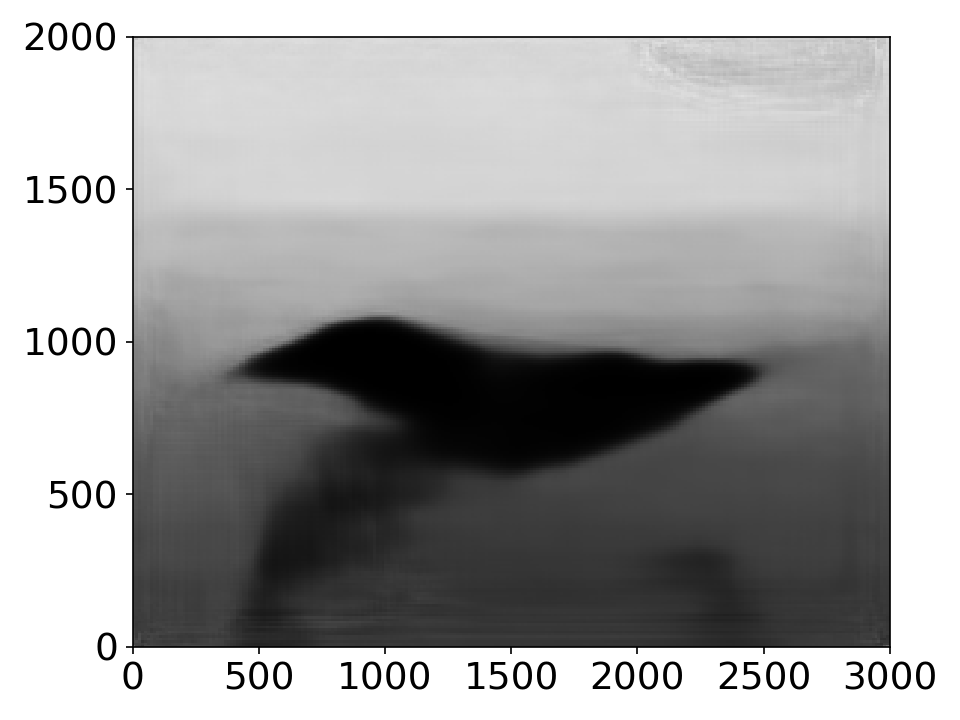}
           \caption{Baseline single model}
        \end{subfigure}
        \begin{subfigure}[b]{0.5\textwidth}
            \centering
            \includegraphics[width=\textwidth]{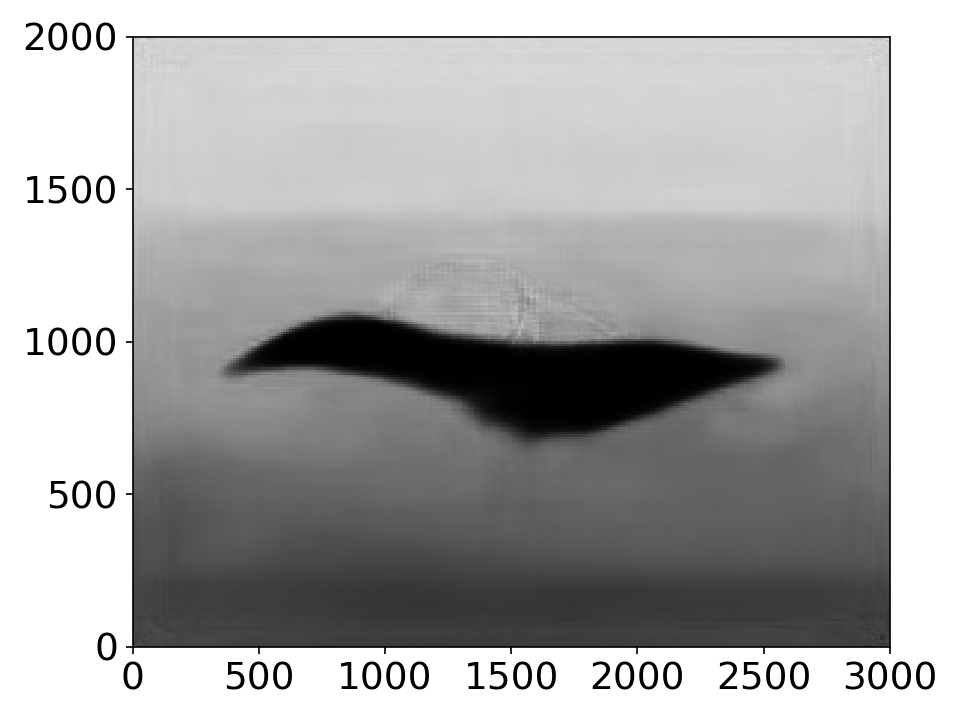}
            \caption{Single model with Fourier images}
        \end{subfigure}
    \end{subfigure}
    \begin{subfigure}[b]{1.0\textwidth}
        \begin{subfigure}[b]{0.5\textwidth}
           \centering
           \includegraphics[width=\textwidth]{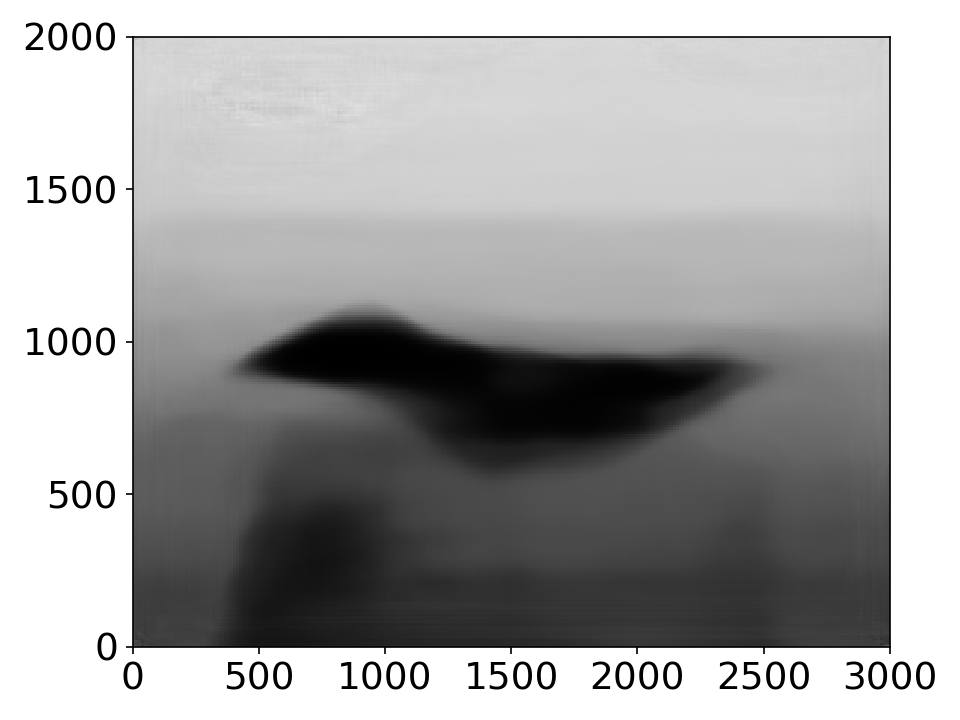}
           \caption{Ensembled model}
        \end{subfigure}
        \begin{subfigure}[b]{0.5\textwidth}
            \centering
            \includegraphics[width=\textwidth]{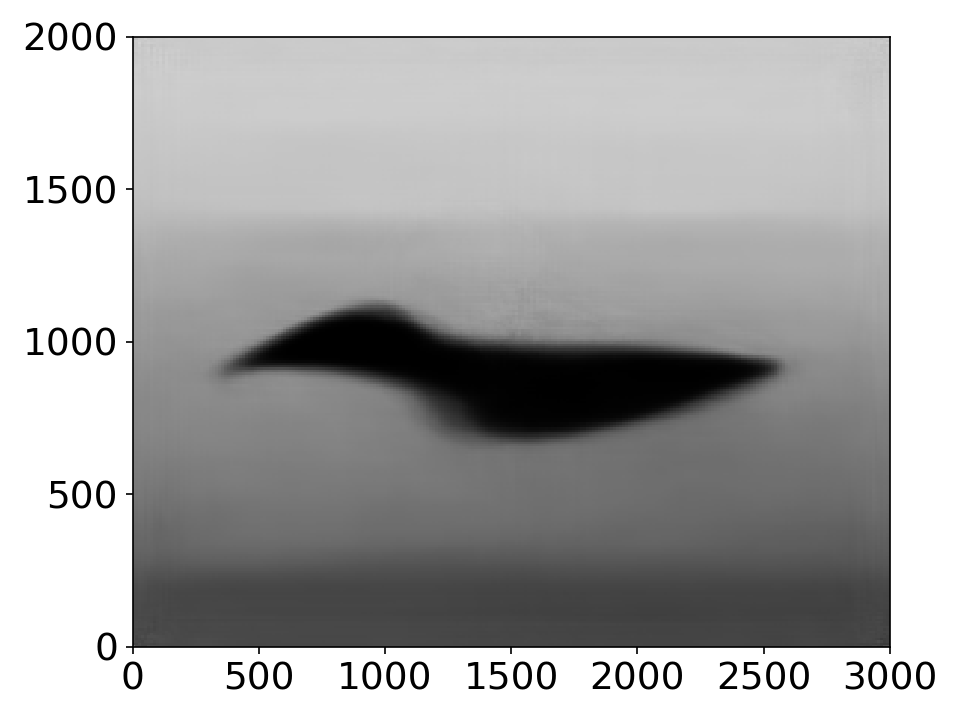}
            \caption{Ensembled model with Fourier images}
        \end{subfigure}
    \end{subfigure}
    \caption{Prediction results for different network architectures}
    \label{fig:prediction_results_2d}
\end{figure}

Figure \ref{fig:prediction_results_2d} presents the prediction results for the ground truth velocity distribution shown on Figure \ref{fig:sample_velocity_model_2}. One can see that adding Fourier images reduces artifacts for both single model instances and ensembles.

\begin{figure}[!h]
    \begin{subfigure}[b]{1.0\textwidth}
        \begin{subfigure}[b]{0.3\textwidth}
           \centering
           \includegraphics[width=\textwidth]{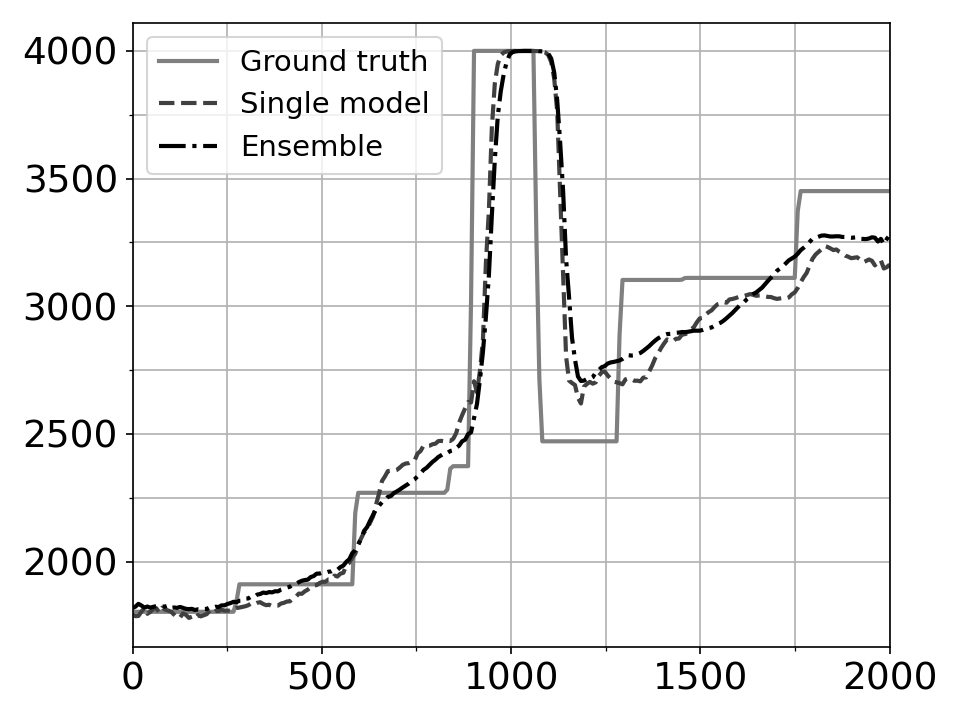}
           \caption{\(x = 750m\)}
        \end{subfigure}
        \begin{subfigure}[b]{0.3\textwidth}
            \centering
            \includegraphics[width=\textwidth]{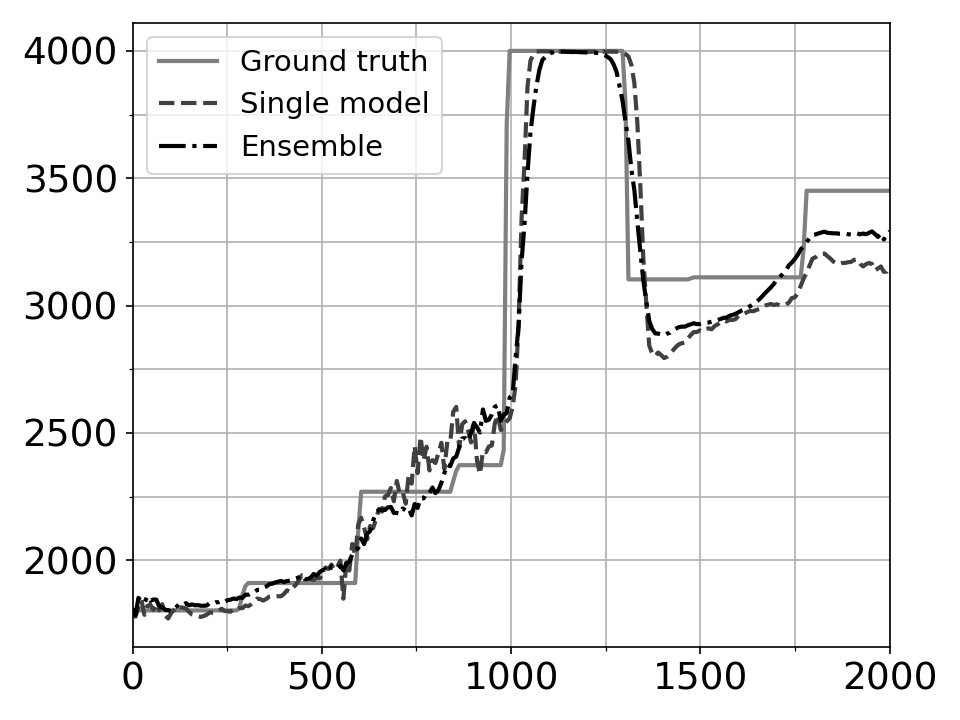}
            \caption{\(x = 1500m\)}
        \end{subfigure}
        \begin{subfigure}[b]{0.3\textwidth}
            \centering
            \includegraphics[width=\textwidth]{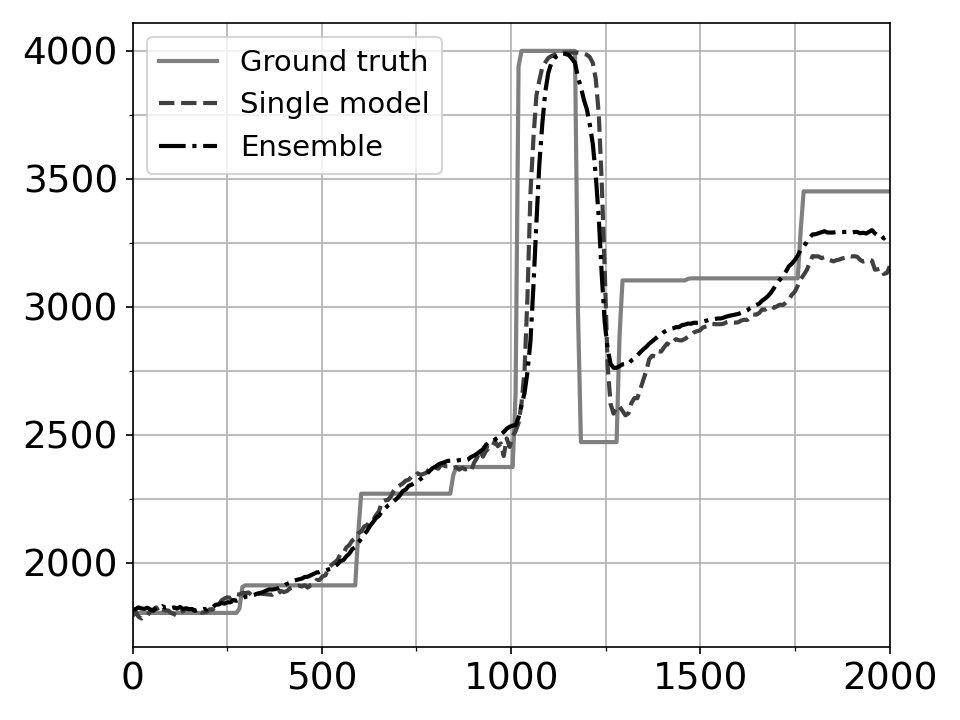}
            \caption{\(x = 2250m\)}
        \end{subfigure}
    \end{subfigure}
    \caption{Vertical profiles of predicted velocity distributions}
    \label{fig:slices_x_single_model_vs_ensemble}
\end{figure}

The difference between single model instances and ensembles can be noted from 1D vertical velocity distribution profiles. Figure \ref{fig:slices_x_single_model_vs_ensemble} presents such profiles along different cross-sections of the same sample depicted on Figure \ref{fig:prediction_results_2d}.
These profiles reveal how the prediction of an ensemble of models differs from a single model participating in it. Ensemble predictions are closer to the ground truth values, and this effect becomes more prominent with the depth (on the right parts of the images). They also tend to be more smooth than a single model's predictions, which is expected due to the aggregating nature of ensembles.

\subsection{Discussion}

The presented method of velocity model acquisition from the boundary measurements demonstrated a good solution quality and can work in real-time. The resulting architecture defined based on this study includes two significant modifications compared with the baseline architecture \cite{Yang2019}: adding Fourier images of the seismograms and using ensembling. These changes allowed us to reach an SSIM value of 0.93 compared with 0.4 mentioned in the previous work.

It was demonstrated that adding Fourier images of raw measured seismic response improves the predicted acoustic impedance distribution quality. This statement passes statistical tests regarding its significance for SSIM metric value. Also, one can see that this modification qualitatively leads to cleaner velocity profiles and removes shady artifacts from the main object.

The regularization approach based on the Sobel filter and implemented as described in the present work was statistically insignificant. Another regularization approach was based on the ensembling technique and was proven to increase the quality of the neural network predictions. Ensembling gives a slight increase in traditional SSIM metric value but provides qualitatively smoother acoustic impedance profiles and better quantitative correspondence with the ground truth properties of the media.

\section{Conclusion}

This work targeted the problem of velocity model acquisition for complex media using convolutional neural networks and based exclusively on boundary measurements. We considered the variation of this problem that is typical for seismic exploration. An open-source dataset of velocity distributions was used to directly compare our results with the results from the previous works. Forward modeling was performed using the grid-characteristic numerical method, the dataset with the numerically obtained seismic records was made publicly available to empower further research in this area.

The modifications of the neural network model proposed in this work allowed us to achieve a structural similarity index measure of $0.93\pm0.01$, which is significantly better than the baseline result. We demonstrated that the changes presented contribute not only to the structural similarity but also to the quantitative correspondence with the ground truth.

The approach, pipeline, and network architecture presented in this work are not specific for seismic exploration and can be transferred to similar areas, such as non-destructive testing and biomedical engineering. Experimenting with these areas is subject to future works.

\section*{Acknowledgements}

AV and AS are supported by RFBR project 18-29-02127 for their work on establishing a numerical pipeline and creating the dataset.
LG is supported by the Basic Research Program at the National Research University Higher School of Economics for his work on forward-modeling, and inverse problem approaches. AU is supported by the Russian Science Foundation under grant agreement 19-71-30020 for his work on ablation study.

\bibliographystyle{unsrt}
\bibliography{main}

\begin{thebibliography}{10}

\bibitem{Ye2018}
Jiaxing Ye, Shunya Ito, and Nobuyuki Toyama.
\newblock Computerized ultrasonic imaging inspection: From shallow to deep
  learning.
\newblock {\em Sensors}, 18(11), 2018.

\bibitem{Tripathi2019}
Gaurav Tripathi, Habib Anowarul, Krishna Agarwal, and Dilip~K. Prasad.
\newblock Classification of micro-damage in piezoelectric ceramics using
  machine learning of ultrasound signals.
\newblock {\em Sensors}, 19(19), 2019.

\bibitem{Perdios2020}
Dimitris Perdios, Manuel Vonlanthen, Florian Martinez, Marcel Arditi, and
  Jean-Philippe Thiran.
\newblock Single-shot cnn-based ultrasound imaging with sparse linear arrays.
\newblock In {\em 2020 IEEE International Ultrasonics Symposium (IUS)}, pages
  1--4, 2020.

\bibitem{Stankevich2021}
Andrey~S. Stankevich, Igor~B. Petrov, and Alexey~V. Vasyukov.
\newblock Numerical solution of inverse problems of wave dynamics in
  heterogeneous media with convolutional neural networks.
\newblock In Margarita~N. Favorskaya, Alena~V. Favorskaya, Igor~B. Petrov, and
  Lakhmi~C. Jain, editors, {\em Smart Modelling for Engineering Systems}, pages
  235--246, Singapore, 2021. Springer Singapore.

\bibitem{PATEL2019}
Dhruv Patel, Raghav Tibrewala, Adriana Vega, Li~Dong, Nicholas Hugenberg, and
  Assad~A. Oberai.
\newblock Circumventing the solution of inverse problems in mechanics through
  deep learning: Application to elasticity imaging.
\newblock {\em Computer Methods in Applied Mechanics and Engineering},
  353:448--466, 2019.

\bibitem{Waldeland2017}
A.U. {Waldeland} and A.H.S.S. {Solberg}.
\newblock Salt classification using deep learning.
\newblock In {\em 79th EAGE Conference and Exhibition 2017}, volume 2017, pages
  1--5, 2017.

\bibitem{Waldeland2018}
Anders~U. {Waldeland}, Are~Charles {Jensen}, Leiv-J. {Gelius}, and Anne
  H.~Schistad {Solberg}.
\newblock Convolutional neural networks for automated seismic interpretation.
\newblock {\em Geophysics}, 37(7):529--537, 2018.

\bibitem{shi2018}
Yunzhi Shi, Xinming Wu, and Sergey Fomel.
\newblock {\em Automatic salt-body classification using deep-convolutional
  neural network}, pages 1971--1975.
\newblock 2018.

\bibitem{zhao2018}
Tao Zhao.
\newblock {\em Seismic facies classification using different deep convolutional
  neural networks}, pages 2046--2050.
\newblock 2018.

\bibitem{Alaudah2019}
Yazeed Alaudah, Patrycja Michałowicz, Motaz Alfarraj, and Ghassan AlRegib.
\newblock A machine-learning benchmark for facies classification.
\newblock {\em Interpretation}, 7(3):SE175--SE187, 2019.

\bibitem{baroni2019}
Lais Baroni, Reinaldo~Mozart Silva, Rodrigo~S. Ferreira, Daniel Civitarese,
  Daniela Szwarcman, and Emilio~Vital Brazil.
\newblock Penobscot dataset: Fostering machine learning development for seismic
  interpretation, 2019.

\bibitem{Salvaris2020}
M.~Salvaris, M.~Kaznady, V.~Paunic, I.~Karmanov, A.~Bhatia, W.H. Tok, and
  S.~Chikkerur.
\newblock Deepseismic: a deep learning library for seismic interpretation.
\newblock In {\em First EAGE Digitalization Conference and Exhibition
  Proceedings}, volume 2020, pages 1--5, 2020.

\bibitem{Liu2019}
Y.~Liu.
\newblock A comparison of machine learning methods for seismic inversion to
  estimate velocity and density.
\newblock In {\em Geoconvention 2019}, pages 13--17, 2019.

\bibitem{Alfarraj2019}
Motaz Alfarraj and Ghassan AlRegib.
\newblock {\em Semi-supervised learning for acoustic impedance inversion},
  pages 2298--2302.
\newblock 2019.

\bibitem{Wang2020}
Lingling Wang, Delin Meng, Bangyu Wu, and Naihao Liu.
\newblock {\em Seismic inversion via closed-loop fully convolutional residual
  network and transfer learning}, pages 1521--1525.
\newblock 2020.

\bibitem{Yang2019}
Fangshu Yang and Jianwei Ma.
\newblock Deep-learning inversion: A next-generation seismic velocity model
  building method.
\newblock {\em Geophysics}, 84(4):R583--R599, 2019.

\bibitem{Das2019}
Vishal Das, Ahinoam Pollack, Uri Wollner, and Tapan Mukerji.
\newblock Convolutional neural network for seismic impedance inversion.
\newblock {\em Geophysics}, 84(6):R869--R880, 2019.

\bibitem{Dujardin2020}
J.R. Dujardin, G.~Sauvin, and M.~Vanneste.
\newblock Acoustic impedance inversion of high resolution marine seismic data
  with deep neural network.
\newblock In {\em NSG2020 4th Applied Shallow Marine Geophysics Conference
  Proceedings}, volume 2020, pages 1--5, 2020.

\bibitem{Araya-Polo2019}
Mauricio {Araya-Polo}, Stuart {Farris}, and Manuel {Florez}.
\newblock Deep learning-driven velocity model building workflow.
\newblock {\em Geophysics}, 38(11), 2019.

\bibitem{Park2020}
Min~Jun {Park} and Mauricio~D. {Sacchi}.
\newblock Automatic velocity analysis using convolutional neural network and
  transfer learning.
\newblock {\em Geophysics}, 85(1), 2020.

\bibitem{gcm_ac}
K.A. Beklemysheva, A.V. Vasyukov, A.O. Kazakov, and I.B. Petrov.
\newblock rid-characteristic numerical method for low-velocity impact testing
  of fiber-metal laminates.
\newblock {\em Lobachevskii J Math}, 39:874–883, 2018.

\bibitem{gcm_residual}
Katerina Beklemysheva, Vasily Golubev, Igor Petrov, and Alexey Vasyukov.
\newblock Determining effects of impact loading on residual strength of
  fiber-metal laminates with grid-characteristic numerical method.
\newblock {\em Chinese Journal of Aeronautics}, 34(7):1--12, 2021.

\bibitem{gcm_migration}
Vasily~I. Golubev.
\newblock The usage of grid-characteristic method in seismic migration
  problems.
\newblock In Igor~B. Petrov, Alena~V. Favorskaya, Margarita~N. Favorskaya,
  Sergey~S. Simakov, and Lakhmi~C. Jain, editors, {\em Smart Modeling for
  Engineering Systems}, pages 143--155, Cham, 2019. Springer International
  Publishing.

\bibitem{gcm_fractured}
Vasily Golubev, Ilia Nikitin, and Anton Ekimenko.
\newblock Simulation of seismic responses from fractured marmousi2 model.
\newblock {\em AIP Conference Proceedings}, 2312(1):050006, 2020.

\bibitem{gcm_multicomponent}
Vasily Golubev, Alexey Shevchenko, and Igor Petrov.
\newblock Simulation of seismic wave propagation in a multicomponent oil
  deposit model.
\newblock {\em International Journal of Applied Mechanics}, 12(08):2050084,
  2020.

\bibitem{numerical_dataset}
Andrey Stankevich, Ivan Nechepurenko, Alexey Shevchenko, Leonid Gremyachikh,
  Andrey Ustyuzhanin, and Alexey Vasyukov.
\newblock Numerical nine-shot seismo records for 1600 acoustic impedance
  distributions, September 2021.

\bibitem{ronneberger2015unet}
Olaf Ronneberger, Philipp Fischer, and Thomas Brox.
\newblock U-net: Convolutional networks for biomedical image segmentation,
  2015.

\bibitem{sobel_filter}
Nick Kanopoulos, Nagesh Vasanthavada, and Robert~L Baker.
\newblock Design of an image edge detection filter using the sobel operator.
\newblock {\em IEEE Journal of solid-state circuits}, 23(2):358--367, 1988.

\bibitem{kingma2014adam}
Diederik~P Kingma and Jimmy Ba.
\newblock Adam: A method for stochastic optimization.
\newblock {\em arXiv preprint arXiv:1412.6980}, 2014.

\bibitem{SSIM}
Zhou Wang, A.C. Bovik, H.R. Sheikh, and E.P. Simoncelli.
\newblock Image quality assessment: from error visibility to structural
  similarity.
\newblock {\em IEEE Transactions on Image Processing}, 13(4):600--612, 2004.

\bibitem{pytorch}
Adam Paszke, Sam Gross, Francisco Massa, Adam Lerer, James Bradbury, Gregory
  Chanan, Trevor Killeen, Zeming Lin, Natalia Gimelshein, Luca Antiga, Alban
  Desmaison, Andreas Kopf, Edward Yang, Zachary DeVito, Martin Raison, Alykhan
  Tejani, Sasank Chilamkurthy, Benoit Steiner, Lu~Fang, Junjie Bai, and Soumith
  Chintala.
\newblock Pytorch: An imperative style, high-performance deep learning library.
\newblock In H.~Wallach, H.~Larochelle, A.~Beygelzimer, F.~d\textquotesingle
  Alch\'{e}-Buc, E.~Fox, and R.~Garnett, editors, {\em Advances in Neural
  Information Processing Systems 32}, pages 8024--8035. Curran Associates,
  Inc., 2019.

\bibitem{shapiro_wilk_statistical_test}
S.~S. Shapiro and M.~B. Wilk.
\newblock {An analysis of variance test for normality (complete samples)†}.
\newblock {\em Biometrika}, 52(3-4):591--611, 12 1965.

\bibitem{ANOVA}
Henry Scheffe.
\newblock {\em The analysis of variance}, volume~72.
\newblock John Wiley \& Sons, 1999.

\end{thebibliography}

\end{document}